\documentclass[letterpaper, 10 pt, conference]{ieeeconf}  
\usepackage{graphicx}
\usepackage{booktabs}
\usepackage{times}
\usepackage{soul}
\usepackage{color}
\usepackage{url}
\usepackage{amsfonts,amssymb}
\makeatletter
\let\NAT@parse\undefined
\makeatother
\usepackage[colorlinks,linkcolor=blue,citecolor=blue]{hyperref}
\usepackage[utf8]{inputenc}
\usepackage[small]{caption}
\usepackage{graphicx}
\usepackage{amsmath}

\usepackage{amsthm}
\usepackage{booktabs}
\usepackage{algorithm}
\usepackage{algorithmic}
\usepackage[switch]{lineno}
\usepackage{booktabs}
 \usepackage{multirow}
 \usepackage{pifont}
 \usepackage{bm}
 \usepackage{tabularx}
 \usepackage{adjustbox}
 \usepackage{makecell}
\usepackage{array}  
\usepackage{caption}
\usepackage{cite}
\renewcommand{\eqref}[1]{Eq.~(\ref{#1})}
\newcommand{\figref}[1]{Fig.~\ref{#1}}

\definecolor{iccvblue}{rgb}{0.21,0.49,0.74}
\definecolor{Light}{rgb}{0.99, 0.92, 0.95}

\usepackage{multirow}
\usepackage[table,xcdraw]{xcolor}
\usepackage{tabularx} 
\usepackage{float}
            
\IEEEoverridecommandlockouts                              

\newcommand{\IEEEauthorrefmark}[1]{\textsuperscript{\small#1}}

\title{\LARGE \bf
Leveraging Learning Bias for Noisy Anomaly Detection*
}

\author{Yuxin Zhang\IEEEauthorrefmark{1},~\IEEEmembership{Graduate Student Member, IEEE}, Yunkang Cao\IEEEauthorrefmark{1},~\IEEEmembership{Graduate Student Member, IEEE}, \\Yuqi Cheng\IEEEauthorrefmark{1},~\IEEEmembership{Graduate Student Member, IEEE}, Yihan Sun\IEEEauthorrefmark{1},~\IEEEmembership{Graduate Student Member, IEEE}, \\Weiming Shen\IEEEauthorrefmark{1},~\IEEEmembership{Fellow,~IEEE}
\thanks{*This work was supported in part by Ministry of Industry and Information Technology of the People's Republic of China under Grant \#2023ZY01089. (\textit{Corresponding author: Weiming Shen.})}
\thanks{$^{1}$Yuxin Zhang, Yunkang Cao, Yuqi Cheng, Yihan Sun, and Weiming Shen are with the State Key Laboratory of Intelligent Manufacturing Equipment and Technology, Huazhong University of Science and Technology, Wuhan 430074, China (e-mail: zyx\_hust@hust.edu.cn; caoyunkang@ieee.org;  yuqicheng@hust.edu.cn; yihansun@hust.edu.cn; wshen@ieee.org). }
}

\begin{document}

\maketitle

\thispagestyle{empty}

\pagestyle{empty}

\begin{abstract}
This paper addresses the challenge of fully unsupervised image anomaly detection (FUIAD), where training data may contain unlabeled anomalies. Conventional methods assume anomaly-free training data, but real-world contamination leads models to absorb anomalies as normal, degrading detection performance. To mitigate this, we propose a two-stage framework that systematically exploits inherent learning bias in models. The learning bias stems from: (1) the statistical dominance of normal samples, driving models to prioritize learning stable normal patterns over sparse anomalies, and (2) feature-space divergence, where normal data exhibit high intra-class consistency while anomalies display high diversity, leading to unstable model responses. Leveraging the learning bias, stage 1 partitions the training set into subsets, trains sub-models, and aggregates cross-model anomaly scores to filter a purified dataset. Stage 2 trains the final detector on this dataset. Experiments on the Real-IAD benchmark demonstrate superior anomaly detection and localization performance under different noise conditions. Ablation studies further validate the framework's contamination resilience, emphasizing the critical role of learning bias exploitation. The model-agnostic design ensures compatibility with diverse unsupervised backbones, offering a practical solution for real-world scenarios with imperfect training data. Code is available at \url{https://github.com/hustzhangyuxin/LLBNAD}.


\end{abstract}



\section{Introduction}
In recent years, unsupervised image anomaly detection (UIAD) methods have garnered increasing attention due to their superior adaptability to long-tailed data distributions compared to supervised approaches~\cite{SAA,GLFM}. Conventional UIAD methods \cite{guo2024dinomaly, luo2025exploring, liu2023simplenet} rely on a strict assumption: the training data must be anomaly-free. If anomalous samples contaminate the training set, the learned feature space may incorrectly incorporate abnormal patterns, reducing detection performance when similar anomalies appear in test.




\begin{figure}[htbp]
\centering
\includegraphics[width=1\linewidth]{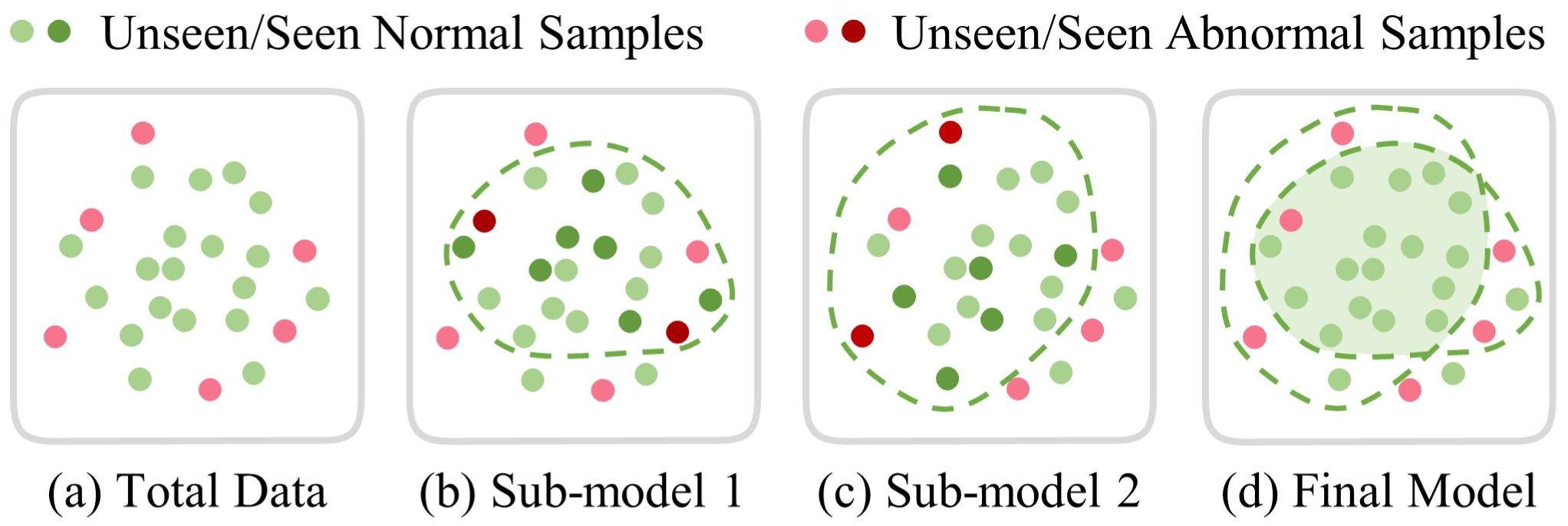}
\caption{Multiple Subset Integration for Robust Anomaly Detection. (a) Overall data distribution with concentrated normal samples (green) and scattered anomalies (red). (b)-(c) Dark red and dark green points denote anomalous and normal samples within the subsets used to train individual sub-models. Green dashed boundaries enclose samples assigned low anomaly scores by each sub-model when evaluated across the entire training set. (d) The green-shaded region identifies high-confidence normal samples consistently achieving low anomaly scores across all sub-models.}
\vspace{-5mm}
\label{fig: motivation}
\end{figure}

However, such a strict anomaly-free requirement is often impractical in real world scenarios~\cite{Survey}, where obtaining perfectly curated normal datasets proves challenging due to inevitable human annotation errors. To address this practical challenge, the formulation of fully unsupervised image anomaly detection (FUIAD) has emerged, which operates under more realistic conditions where training data may contain unlabeled anomalies. The core difficulty of FUIAD lies in preventing the model from inadvertently learning abnormal features during training, which would diminish the discriminative power between normal and abnormal patterns during inference. Therefore, the key to solving the FUIAD problem lies in avoiding noise as much as possible. Among existing FUIAD methods, SoftPatch \cite{jiang2022softpatch} uses a noise discriminator to filter out noisy data before coreset construction, and softens the search process to reduce the impact of hard unfiltered noisy samples. However, discriminators may overfit to limited anomaly types while potentially eliminating critical normal features. FUN-AD \cite{Im_2025_WACV} assigns pseudo-labels based on iteratively re-constructed memory bank and pairwise-distance statistics to achieve robustness to initial noisy labels. But inaccurate pseudo-labels may mislead the training direction of the model, causing the model to learn incorrect patterns.  InReaCh \cite{InReaCh} extracts high-confidence nominal patches from training data by associating them between image realizations into channels, taking into account only channels with high spans and low spread.  However, in complex real-world scenarios, abnormal data may also exhibit similar statistical characteristics in these channels, leading to misclassifying some abnormal data as normal, thus reducing the accuracy of anomaly detection. 






To better address this noise avoidance challenge, we conduct a systematic analysis of feature learning mechanisms in FUIAD frameworks. We identify distinct disparities in how models acquire normal versus anomalous features, and this inherent learning bias provides a principled foundation for our method design. First, models manifest faster acquisition rates for dominant features compared to rare features. In FUIAD tasks, the scarcity of anomalous samples coupled with a substantially smaller proportion of anomalous pixels relative to normal pixels creates an imbalanced learning scenario. This imbalance leads to accelerated learning of normal patterns (dominant features) while impeding the effective acquisition of anomalous characteristics (rare features). Second, normal samples demonstrate low intra-class variation in feature space, whereas anomalous samples exhibit a high diversity of defect patterns. The inherent feature consistency in normal samples enables strong generalization capabilities for novel normal instances, while the intrinsic heterogeneity of anomalies results in a significant variance in detection performance when encountering unseen anomaly types.

Building upon this insight into inherent learning bias, we propose to strategically partition the training set into multiple subsets and train specialized sub-models on each. This design capitalizes on the observation that sub-models exposed to partial feature spaces exhibit polarized detection behaviors: their constrained training scope maintains robust confidence in prevalent normal patterns, yet induces heightened inter-model disagreement when encountering rare anomalies across subsets. As shown in \figref{fig: motivation} (a), normal samples form compact distributions in feature space due to their low intra-class variation, while anomalies exhibit dispersed patterns resulting from high defect diversity. To exploit this disparity, the training set is randomly divided into multiple subsets (e.g., the dark red and dark green points in \figref{fig: motivation} (b)-(c)), and a corresponding sub-model is trained for each subset. When tested on the entire training set, different sub-models exhibit varying performance. During testing on the entire training set, samples within the green dashed line in \figref{fig: motivation} (b)-(c) indicate low anomaly scores assigned by the sub-model. Due to the learning bias, models generalize well to unseen normal patterns (dominant features with low intra-class similarity) but show higher anomaly scores for unseen anomalous patterns (rare features with high diversity). By aggregating scores across all sub-models, we identify high-confidence normal samples as those consistently residing in low-score regions (green-shaded area in \figref{fig: motivation} (d)).


This study makes the following contributions to address training set contamination in anomaly detection. We first establish through systematic analysis that direct applications of existing unsupervised methods face fundamental limitations due to contaminated training data, identifying contamination reduction as the core challenge. To solve this, we develop a two-stage framework integrating contamination-resistant screening and anomaly detection, where the first stage employs multiple subset generation and cross-model consensus to effectively purify the training data. Furthermore, our approach transforms neural networks' inherent limitation in low-frequency feature capture into an advantage by leveraging their natural learning bias as a noise filter, providing new perspectives for handling contaminated training scenarios.

\section{Problem Definition}
FUIAD aims to learn normal patterns from contaminated datasets and identify deviations from learned normal patterns during inference. Formally, given a training set $ \mathcal{D}_{\text{train}} = \{x_i\}_{i=1}^N $ containing a noise ratio $\alpha  \in [0, 1)$, where $\alpha $ denotes the proportion of anomalous samples. The task comprises two core objectives: during training, mitigate contamination interference through robust learning strategies to establish accurate normal pattern characterization; during inference, utilize an anomaly scoring function  $g: p \to \mathbb{R}$  to assign higher scores to pixels deviating from the learned normal patterns to achieve anomaly localization.

The noise ratio $\alpha $ impacts models' robustness. Low $\alpha $ preserves discriminability of normal patterns, while high $\alpha $ degrades performance by conflating normal and anomalous features. This necessitates reducing training set contamination.To mitigate this, we introduce a contamination-robust denoising framework that leverages models' inherent learning bias to identify and filter anomalous samples, thereby reducing contamination. Our method enables seamless integration with state-of-the-art UIAD backbones, adapting them to FUIAD scenarios via contamination-robust preprocessing.
\section{Learning Bias in Anomaly Detection Models}
Learning bias refers to the systematic discrepancy in a model's capacity to assimilate and distinguish normal versus anomalous patterns, originating from the dual origins of (1) the statistical dominance of normal samples during training, and (2) the intrinsic divergence in feature space distributions between classes.

The statistical dominance mechanism fundamentally shapes model behavior through imbalanced gradient updates. When exposed to feature categories with imbalanced population distributions, models demonstrate accelerated assimilation and enhanced memorization toward dominant features through positive feedback in parameter updates. In industrial anomaly detection contexts, this bias is exacerbated by two intrinsic data characteristics: (1) the scarcity of anomalous samples (often $<5\%$ of training data) and (2) the limited spatial prevalence of anomalies (typically occupying $<1\%$ of image pixels). These constraints drive models to prioritize stable normal representations (e.g., regular textures and structural configurations of industrial products) through repeated exposure, while their capacity to capture anomalous features remains underdeveloped due to insufficient exposure frequency and limited spatial activation. 

Essential differences exist in the feature space distributions between normal and anomalous samples. Normal samples exhibit low intra-class variation as a direct consequence of manufacturing precision controls. This feature-space consistency enables models to achieve strong generalization performance on unseen normal instances. Conversely, industrial anomalies span diverse subtypes (e.g., mechanical damage, material degradation, assembly errors) with visually distinct characteristics and non-stationary feature distributions. This intrinsic heterogeneity leads to significant performance variance when models encounter anomaly types absent from training data, particularly evident in open-world cross-category detection scenarios.


\section{Method}

\begin{figure*}[htbp]
\centering
\includegraphics[width=\linewidth]{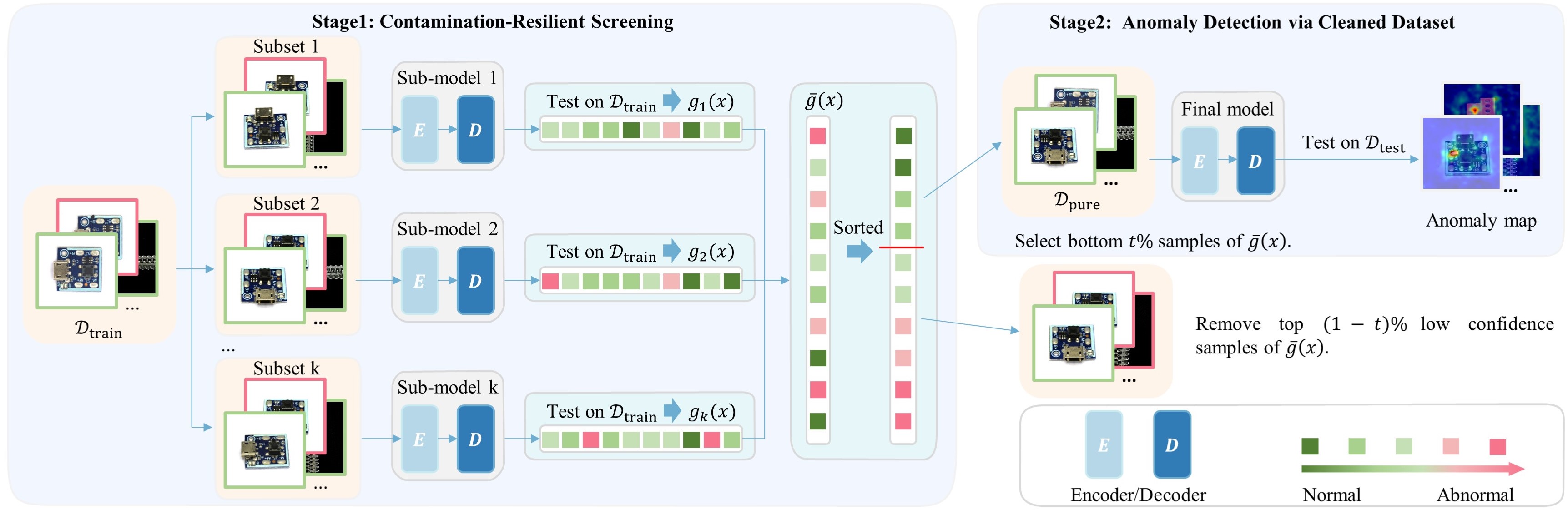}
\caption{Framework of the proposed method. Our framework comprises two stages. In stage 1, $\mathcal{D}_{\text{train}}$ is partitioned into $k$ subsets for individual model training. Each model is then tested on the whole training set to yield anomaly scores. The scores are aggregated into consensus anomaly scores, sorted, and the t\% of samples with the lowest scores are selected as $\mathcal{D}_{\text{pure}}$. In stage 2, the final model is trained on $\mathcal{D}_{\text{pure}}$ for anomaly detection.}
\vspace{-5mm}
\label{fig:framework}
\end{figure*}

Capitalizing on the distinct feature learning mechanisms identified in our analysis, we develop a two-stage framework, as shown in \figref{fig:framework}.

Stage 1: Contamination-Resistant Screening. We leverage inherent feature learning bias in models' responses to known and unknown features to achieve the separation of normal and abnormal samples. 
Despite training on mixed normal and anomalous samples, UIAD models exhibit distinct learning efficacy for normal versus anomalous features. 
Specifically, models generalize well to unseen normal patterns but produce significant reconstruction errors for unseen anomalies due to their sparse and divergent feature distributions.
To exploit this bias, we partition the training set into multiple subsets and train dedicated sub-models on each. During cross-subset evaluation, sub-models assign low anomaly scores to normal samples from other subsets while yielding higher scores for anomalies. By aggregating these responses, we effectively separate normal and anomalous samples.

Stage 2: Anomaly Detection via Cleaned Dataset. Using the cleaned dataset obtained in stage 1, we train a final model for anomaly detection. Suppressed from anomalous interference, this model achieves enhanced discriminative power by focusing exclusively on normal feature manifolds.

\subsection{Contamination-Resilient Screening}
Exploiting the intrinsic learning bias where normal patterns exhibit stable representations while anomalies manifest inconsistent features, our framework separates contamination through their differential behavior across multiple sub-models. By training multiple sub-models on partitioned subsets of $\mathcal{D}_{\text{train}}$, we observe that normal samples consistently yield low anomaly scores across all sub-models due to their uniform feature distributions, while anomalies exhibit unstable anomaly scores from their sparse and diverse patterns. Aggregated anomaly scores enable contamination filtering via quantile thresholding, yielding $\mathcal{D}_{\text{pure}}$ with suppressed noise for subsequent detection.
\subsubsection{Multiple Subset Generation}

The training set  $\mathcal{D}_{\text{train}} = \{x_i\}_{i=1}^N$  is  randomly partitioned into $k$ subsets  $\{S_j\}^k_{j=1}$, where $\bigcup_{j=1}^{k} S_j = \mathcal{D}_{\text{train}}$. Each subset $S_j$ is used to train a sub-model $M_j$, and the training objective is to minimize the loss function:

\begin{equation}\label{loss}
\begin{split}
\theta_j^* = \underset{\theta_j}{\arg\min} \sum_{x_i \in S_j} \mathcal{L}(M_j(x_i)), \quad \forall j \in \{1, \ldots, k\}
\end{split}
\end{equation}
\noindent where \(\theta_j\) is the parameter of \(M_j\).



For any sample $x_i \in S_j$, the cross-model divergence between its native sub-model $M_j$ and non-native sub-models $\{M_l\}_{l \neq j}$ can be defined as:

\begin{equation}\label{Delta_x}
\begin{split}
\Delta(x_i) = \mathbb{E}_{l \neq j} \left[ \left\| f_{\theta_j}(x_i) - f_{\theta_l}(x_i) \right\|_2 \right]
\end{split}
\end{equation}

For a normal sample, sub-models trained on different subsets should learn similar normal patterns. Thus, their feature representations $f_{\theta_j}(x_i)$ and $f_{\theta_l}(x_i)$ will be consistent, leading to a small $\Delta(x_i)$. For an anomalous sample, its native sub-model $M_j$ may erroneously assimilate $x_i$'s patterns due to exposure during training, while non-native sub-models $\{M_l\}_{l \neq j}$ (trained on disjoint subsets) maintain unbiased feature representations. This introduces a consensus discrepancy between models and results in a large $\Delta(x_i)$.

\subsubsection{Cross-Model Consensus Purity Evaluation}
By systematically aggregating the responses of the sub-models to all training samples, we establish an anomaly likelihood metric. This metric can effectively distinguish between normal and abnormal samples, resulting in a dataset that contains either a small number of or no abnormal samples.

For each sub-model $M_j$ trained on subset $S_j$, the learned feature representations exhibit asymmetric behavior between normal and anomalous samples. Due to the dominance of normal features in $S_j$, $M_j$ establishes robust representations for these patterns. Specifically, coupled with their low intra-class variation, normal samples tend to exhibit consistently low anomaly scores $g(x_i)$ between $x_i \in S_j$ and $x_i \notin S_j$, as reflected by minimal feature deviations and negligible cross-model divergence ($\Delta(x_i) \rightarrow 0$, \eqref{Delta_x}). In contrast, anomalies, despite potential partial assimilation during training, still remain poorly learned due to their sparse representation and high diversity of defect patterns. This heterogeneity results in relatively higher anomaly scores for anomalies within the subset $S_j$ and even higher anomaly scores for cross-subset anomalies ($x_i \notin S_j$).

Given these observations, it is essential to aggregate the anomaly scores across all sub-models to obtain a comprehensive assessment of each sample's anomaly status. For sub-model $M_j$, the anomaly scores for all samples can be denoted as $g_j(x_i)$, where $\forall x_i \in \mathcal{D}_{\text{train}}$. For each sample $x_i$, aggregate the anomaly scores from all $k$ sub-models:

\begin{equation}\label{g}
\begin{split}
\bar{g}(x_i) = \frac{1}{k} \sum_{j=1}^{k} g_j(x_i)
\end{split}
\end{equation}

Based on consensus anomaly scores $\{\bar{g}(x_i)\}_{i=1}^{N}$, we retain the samples in the lowest $t$-fraction of the distribution:

\begin{equation}\label{}
\begin{split}
\mathcal{D}_{\text{pure}} = \left\{ x_i \in \mathcal{D}_{\text{train}} \,\middle|\, \bar{g}(x_i) \leq \tau_t \right\}
\end{split}
\end{equation}

\noindent where the threshold $\tau_t$ is defined as the $t$-th quantile of the score distribution:
\begin{equation}\label{threshold}
\begin{split}
\tau_t = \text{Quantile}\left(\left\{\bar{g}(x_i)\right\}_{i=1}^{N}, t\right)
\end{split}
\end{equation}

 Applying the quantile-based threshold $\tau_t$ to the aggregated anomaly scores, we obtain a high purity data set $\mathcal{D}_{\text{pure}}$, in which potential anomalies are effectively suppressed. $\mathcal{D}_{\text{pure}}$ serves as the foundation for subsequent model training.

\subsection{Anomaly Detection via Cleaned Dataset}
In our contamination-resilient screening stage, we first partition the original training set $\mathcal{D}_{\text{train}}$ into $k$ disjoint subsets $\{S_k\}^k_{k=1}$. For each subset $S_j$, we train a dedicated sub-model $M_j$ that learns distinctive feature representations. By systematically aggregating anomaly scores across all sub-models via \eqref{g}, we establish cross-model consensus metrics to derive the cleaned dataset $\mathcal{D}_{\text{pure}}$ through quantile thresholding $\tau_t$. Subsequently, this cleaned dataset $\mathcal{D}_{\text{pure}}$ will be utilized to train a final detection model.

Feature learning bias exists in arbitrary UIAD models; thus, the sub-models and our final model can also be arbitrary UIAD models. In this paper, we have selected Dinomaly \cite{guo2024dinomaly} for preliminary investigation.



Dinomaly is a minimalist reconstruction-based anomaly detection framework built exclusively on pure Transformer architectures. It consists of an encoder, a bottleneck, and a reconstruction decoder. Let $E$, $B$, and $D$ denote the encoder, bottleneck, and decoder modules respectively, where $F_E = E(I)$ extracts latent features, $F_B = B(F_E)$ processes them, and $\hat{I} = D(F_B)$ reconstructs the output. During training, the decoder is optimized to reconstruct intermediate encoder features by maximizing cosine similarity between feature maps. During inference, the decoder accurately reconstructs normal regions but fails to reconstruct anomalous regions as it lacks exposure to anomaly patterns during optimization. These reconstruction errors are used for anomaly detection, where higher deviations indicate a greater likelihood of anomalies. Under the FUIAD setting, Dinomaly’s inherent learning bias is manifested through its architectural design. Due to limited exposure to anomalous samples during training, the model exhibits amplified reconstruction errors for less frequently encountered anomalies. This systematic discrepancy between normal and anomalous feature reconstructions enables effective anomaly detection despite training data contamination. 




\section{Experiments}
\subsection{Experiment Setting}
\subsubsection{Dataset Description}



\begin{table*}[t]
\centering
\caption{FUIAD performance(I-AUROC/AUPRO) comparisons on Real-IAD \cite{wang2024real} The results are over 30 categories. The best performance is in \textbf{bold}, and the second best is in \underline{underlined.}}
\label{table:FUIAD performance comparisons} 
\fontsize{11}{14}\selectfont{
\resizebox{\linewidth}{!}{
\begin{tabular}{c|cccccccc|>{\columncolor{blue!8}}c|}
\toprule[1.5pt]
\textbf{Settings} & \textbf{PaDim \cite{defard2021padim}} & \textbf{CFlow \cite{gudovskiy2022cflow}} & \textbf{PatchCore \cite{roth2022towards}} & \textbf{SoftPatch \cite{jiang2022softpatch}} & \textbf{SimpleNet \cite{liu2023simplenet}} & \textbf{DeSTSeg \cite{zhang2023destseg}} & \textbf{RD \cite{deng2022anomaly}} & \textbf{UniAD \cite{you2022unified}} & \textbf{Ours} \\ 
\midrule
$\alpha=0.0$           & 84.6/84.4      & 83.9/90.6      & 91.3/92.6          & \underline{91.4}/92.1          & 89.8/83.9          & 89.6/88.7        & 89.3/\underline{95.0}   & 85.4/87.6      & \textbf{93.4}/\textbf{95.9}     \\
$\alpha=0.1$           & 81.9/86.4      & 80.3/90.7      & 90.4/93.2          & \underline{90.9}/92.9          & 83.3/79.9          & 85.6/86.9        & 88.1/\underline{95.1}   & 84.2/87.7      & \textbf{91.5}/\textbf{95.8}     \\
$\alpha=0.2$           & 80.1/86.5      & 79.6/90.7      & 89.5/93.0          & \textbf{90.5}/92.9          & 79.6/75.9          & 80.3/83.2        & 87.3/\underline{94.9}   & 82.8/87.3      & \underline{89.8}/\textbf{95.6}     \\
$\alpha=0.4$           & 77.0/86.1      & 78.0/90.2      & \underline{88.1}/92.4          & \textbf{89.3}/92.5          & 74.7/70.5          & 74.4/75.5        & 84.5/\underline{94.7}   & 80.1/86.6      & 87.0/\textbf{94.8}     \\
\bottomrule[1.5pt]
\end{tabular}}}
\end{table*}

\begin{figure*}[htbp]
\centering
\includegraphics[width=\linewidth]{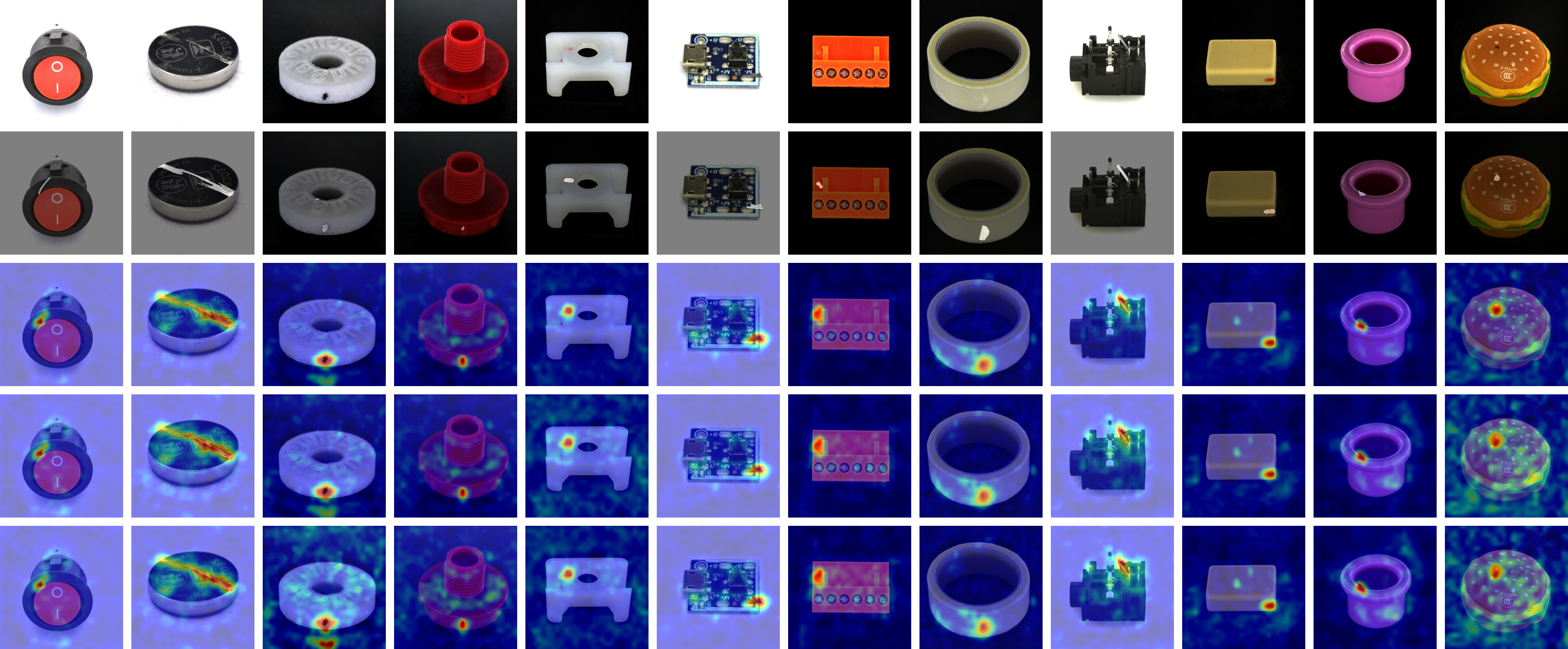}
\caption{Qualitative results of the proposed method. From top to bottom: the input image, the ground truth masks, and the output anomaly maps of noise ratio $\alpha=0.1, 0.2, 0.4$.}
\vspace{-5mm}
\label{fig: Anomaly_maps_visualization}
\end{figure*}

\begin{figure*}[htbp]
\centering
\includegraphics[width=0.8\linewidth]{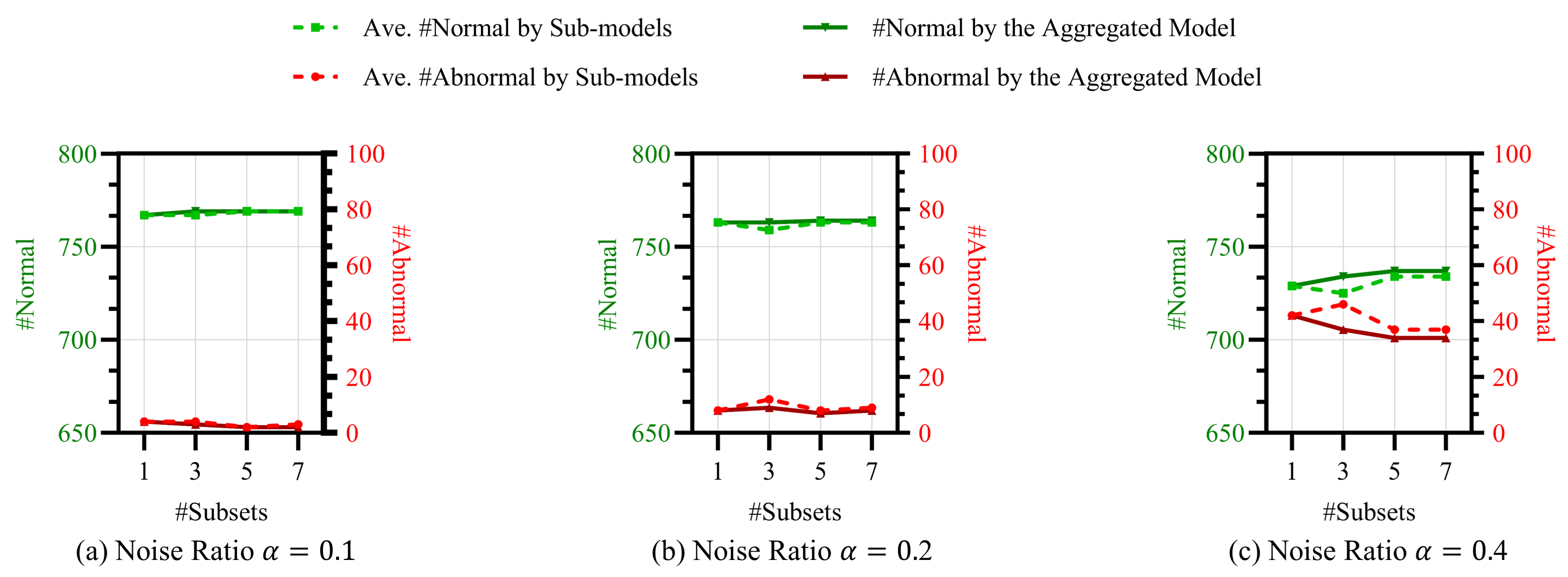}
\caption{Normal vs. Abnormal Samples in Sub-models and Aggregated Models under Different Noise Ratios. \#Normal and \#Anomaly denote the number of selected normal and anomaly samples, respectively.}
\vspace{-5mm}
\label{fig:line_graph}
\end{figure*}

Real-IAD \cite{wang2024real} constitutes a large-scale, real-world, multi-view industrial anomaly detection benchmark comprising 150,000 images across 30 distinct industrial objects. Under the FUIAD setting, each object category in the test set contains 500 normal and 500 anomalous images. The training sets maintain constant per-category sample counts, with anomaly contamination levels controlled by specified noise ratios. Distinguished by its substantial number of anomalous samples and larger defect proportions compared to existing benchmarks, Real-IAD facilitates the training of robust anomaly detection models while enabling fair performance evaluation.

\subsubsection{Evaluation Metrics}
At the image level, detection performance is quantified using the Area Under the Receiver Operating Characteristic curve (I-AUROC). For pixel-level anomaly localization, we employ the normalized Area Under the Per-region Overlap curve (AUPRO). Both metrics yield values within [0,1], where higher values indicate superior detection capability.

\subsubsection{Implementation Details}
We evaluate the noise ratios $\alpha \in \{0, 0.1, 0.2, 0.4\}$ with $\alpha=0$ as the clean upper bound. The training set $\mathcal{D}_{\text{train}}$ is split into subsets $k \in \{1,3,5,7\}$, with $k=5$ by default, each trained for 1000 iterations. Using Dinomaly \cite{guo2024dinomaly} as the core framework for training models in both stage 1 and stage 2, we employ DINOv2-R pre-trained ViT-Base/14 (patch size = 14) with register tokens as the backbone. Noisy Bottleneck applies 0.2 dropout. Input images are resized to $448\times448$ and center-cropped to $392\times392$, yielding $28\times28$ feature maps. In \eqref{loss}, reconstruction employs 2-group loose constraints and the hard-mining global cosine loss, which detaches the gradients of well-reconstructed features to prevent overfitting. Following \eqref{threshold}, we set $t=40\%$ to define the quantile threshold $\tau_t$, retaining the lowest 40\% anomaly scores to generate $\mathcal{D}_{\text{pure}}$. All experiments are conducted on NVIDIA A100 GPUs using PyTorch 2.6.0.


\subsection{Comparisons with State-of-the-art methods}
To demonstrate the superiority of our method, we conduct comparative experiments with the approaches mentioned in the Real-IAD \cite{wang2024real}, including PaDim \cite{defard2021padim}, Cflow \cite{gudovskiy2022cflow}, PatchCore \cite{roth2022towards}, SoftPatch \cite{jiang2022softpatch}, SimpleNet \cite{liu2023simplenet}, DeSTSeg \cite{zhang2023destseg}, RD \cite{deng2022anomaly}, and UniAD \cite{you2022unified}, evaluating their performance under the FUIAD setting. The main results are presented in Table \ref{table:FUIAD performance comparisons}.

\subsubsection{Anomaly Detection}
Our method achieves state-of-the-art (SOTA) performance at $\alpha=0$ and $\alpha=0.1$, and attains the second-best result at 
$\alpha=0.2$. The superior performance under low noise ratios stems from the intrinsic learning bias: the statistical dominance of normal samples in $\mathcal{D}_{\text{train}}$ drives sub-models to establish robust normal feature representations, while sparse anomalies exhibit amplified reconstruction errors due to underrepresented feature distributions. However, as $\alpha\geq 0.2$, the reduced dominance of normal samples weakens this learning bias. Sub-models increasingly assimilate anomalous features from their contaminated subsets, improving anomaly reconstruction capability and
diminishing discriminative power for unseen anomalies during testing. Consequently, the framework's ability to distinguish novel anomalies degrades with higher contamination levels.


\subsubsection{Anomaly Localization}
We compare anomaly localization performance across existing methods on the Real-IAD benchmark. Results show our method consistently achieves SOTA performance under all evaluated noise ratios $\alpha \in \{0, 0.1, 0.2, 0.4\}$. Notably, even under high contamination ($\alpha\geq 0.2$), our method maintains exceptional localization accuracy with AUPRO scores exceeding 94\%, identifying a variety of defect regions. Some qualitative results are shown in \figref{fig: Anomaly_maps_visualization}. It demonstrates superior anomaly localization capability across diverse defect types, including scratches, mechanical deformations, and porosity anomalies.


\subsection{Ablation Studies}
In this subsection, we conduct comprehensive experiments to evaluate the impact of the number of subsets affects model performance under different noise ratios $\alpha$, as well as the effectiveness of the multi-model score aggregation strategy. As shown in \figref{fig:line_graph}, the green and red lines denote the number of selected normal and anomaly samples, respectively. And the dashed and solid lines correspond to the aggregated models and sub-models, respectively.

\subsubsection{Number of Subsets}

As indicated by the red dashed line in \figref{fig:line_graph}, the aggregated model yields $\mathcal{D}_{\text{pure}}$ with fewer anomalous samples across all noise ratios when using subsets $k>1$ compared to the non-partitioned baseline $k=1$. This systematic reduction in contamination levels demonstrates the effectiveness of our subset partitioning strategy for suppressing anomalies in $\mathcal{D}_{\text{pure}}$. The cross-model anomaly score aggregation strategy achieves optimal performance at $k=5$ across all noise ratios ($\alpha \in \{0.1, 0.2, 0.4\}$).  For smaller subset counts $k \in \{1, 3\}$, the higher proportion of anomalies in each subset diminishes the learning bias effect. This causes sub-models to over-adapt to diverse anomalous features, reducing reconstruction errors for anomalies and thereby impairing normal-anomaly discriminability. Conversely, larger subset counts $k=7$ limit the number of normal samples per subset, compromising the model's ability to comprehensively learn normal feature manifolds and degrading generalization performance on unseen normal instances. The $k=5$ configuration optimally balances these opposing factors: it preserves sufficient normal samples per subset to maintain learning bias while controlling anomaly presence to prevent overfitting.

\subsubsection{Effectiveness of Multi-Model Anomaly Score Aggregation}

To validate the effectiveness of cross-model anomaly score aggregation strategy, we analyze the contamination levels in $\mathcal{D}_{\text{pure}}$ derived from two approaches: (1) individual sub-models and (2) our aggregated anomaly scoring method defined in \eqref{g}, which combines outputs from all sub-models before threshold-based $\mathcal{D}_{\text{pure}}$ generation. As shown in \figref{fig:line_graph}, the aggregated model (red dashed line) consistently retains fewer anomalous samples in $\mathcal{D}_{\text{pure}}$ compared to single-submodel baselines (red solid line) across all tested subset counts ($k \in \{3,5,7\}$) and noise ratios ($\alpha \in \{0.1, 0.2, 0.4\}$). This systematic reduction in contamination demonstrates that our aggregation strategy effectively suppresses anomalies by leveraging the models' learning bias.


\section{Conclusion}
The fundamental challenge in fully unsupervised image anomaly detection (FUIAD) lies in preventing models from assimilating anomalous features during training. To address this, this work presents a novel two-stage framework that systematically exploits inherent learning bias to isolate and suppress anomalous patterns.
During stage 1, we partition the dataset into subsets, train submodels on each, and aggregate their anomaly scores to filter out a cleaned training set with significantly reduced contamination. Next, stage 2 trains the final anomaly detector on $\mathcal{D}_{\text{pure}}$ for improved IAD performance. The model-agnostic architecture permits seamless integration with various UIAD backbones. Preliminary investigations using Dinomaly on Real-IAD demonstrate the efficacy of our subset partitioning and cross-model fusion strategy, particularly achieving SOTA anomaly localization performance.
Future work will explore adaptive subset partitioning and advanced fusion mechanisms to further enhance contamination resilience.

\addtolength{\textheight}{-12cm}   






\bibliography{references.bib}

\end{document}